\pdfoutput=1
\documentclass[11pt]{article}
\usepackage[final]{acl}
\usepackage{times}
\usepackage{latexsym}

\usepackage[T1]{fontenc}
\usepackage[utf8]{inputenc}
\usepackage{microtype}
\usepackage{inconsolata}
\usepackage{graphicx}

\usepackage{microtype}
\usepackage{hyperref}
\usepackage{url}
\usepackage{booktabs}

\usepackage[ruled,vlined]{algorithm2e}
\usepackage[utf8]{inputenc}
\usepackage[T1]{fontenc}
\usepackage{CJK}
\usepackage{algorithmic}
\usepackage{multirow}
\usepackage{multicol}
\usepackage{booktabs}
\usepackage{xpinyin}
\usepackage{amsmath}
\usepackage{graphicx}
\usepackage{wrapfig}
\usepackage{adjustbox}
\usepackage{array}
\usepackage{xcolor}
\usepackage{tcolorbox}
\usepackage{makecell}
\usepackage{color}
\usepackage{booktabs}
\usepackage{mdframed}
\usepackage{ruby}
\usepackage{pinyin}

\definecolor{Melon}{RGB}{242, 150, 26}
\newtcolorbox[list inside=prompt,auto counter,number within=section]{prompt}[1][]{
    colbacktitle=black!24,
    coltitle=black,
    fontupper=\normalsize,
    boxsep=3pt,
    left=0pt,
    right=0pt,
    top=0pt,
    bottom=0pt,
    boxrule=1pt,
    #1,
}

\title{Cracking the Code: Enhancing Implicit Hate Speech \\ Detection through Coding Classification}

\author{
Lu Wei${^{1}}$, Liangzhi Li${^{1}}$\thanks{Corresponding author.}, Tong Xiang${^{1}}$, Xiao Liu${^{2}}$, Noa Garcia${^{1}}$ \\
${^{1}}$The University of Osaka, Osaka, Japan \\
${^{2}}$Meetyou AI Lab, Xiamen, China \\
\texttt{\{lu-wei,tongxiang\}@is.ids.osaka-u.ac.jp} \\
\texttt{\{li,noagarcia\}@ids.osaka-u.ac.jp}, \texttt{runnishino@gmail.com}
}

\begin{document}
\maketitle
\begin{abstract}
The internet has become a hotspot for hate speech (HS), threatening societal harmony and individual well-being. While automatic detection methods perform well in identifying explicit hate speech (ex-HS), they struggle with more subtle forms, such as implicit hate speech (im-HS). We tackle this problem by introducing a new taxonomy for im-HS detection, defining six encoding strategies named \textit{codetypes}. We present two methods for integrating codetypes into im-HS detection: 1) prompting large language models (LLMs) directly to classify sentences based on generated responses, and 2) using LLMs as encoders with codetypes embedded during the encoding process. Experiments show that the use of codetypes improves im-HS detection in both Chinese and English datasets, validating the effectiveness of our approach across different languages.

\textcolor{red}{\textbf{NOTE}: The samples presented in this paper may be considered offensive or vulgar.}
\end{abstract}

\section{Introduction}
In the current socio-cultural context, the identification of hate speech (HS) has become increasingly important~\cite{das2020hate, weidinger2022taxonomy, yin2022hidden}. Numerous studies highlight the negative impact of toxic language and HS, not only on the directly targeted individuals~\citep{jikeli2023antisemitic, hettiachchi2023crowd, miller2023identity, klutse2023dismantling, sharma2022detection, xu2022much} but also on the society as a whole~\citep{erjavec2012you, saha2019prevalence, kiritchenko2021confronting, rapp2021social, maarouf2022virality, aleksandric2022twitter}. HS has the potential to exacerbate divisions and conflicts~\citep{schmitz2022quantifying}, and in extreme cases, can threaten community stability~\citep{perez2023nlp, williams2020hate}. 

\begin{figure}[t]
  \includegraphics[width=\columnwidth]{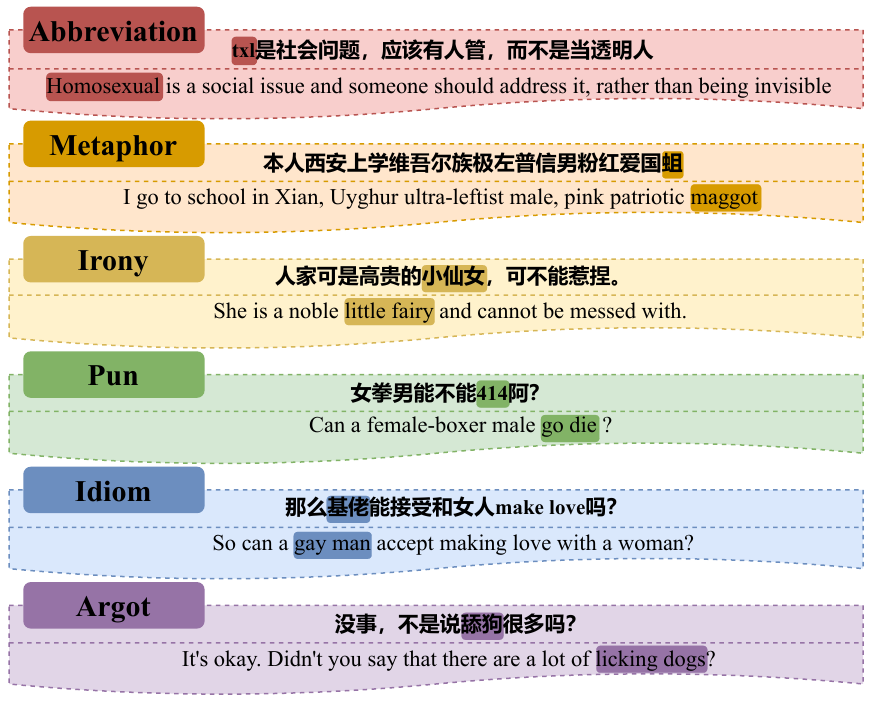}
  \caption{Selected examples from the ToxiCN dataset~\citep{lu2023facilitating} that illustrate six codetypes. English translation below is provided for clarity. The keywords that encode specific types of implicit hate information are highlighted.}
  \label{fig:ideaFramework}
\end{figure}

Within the prevailing research, the majority of scholarly efforts are dedicated to categorizing macro-level concepts of HS~\citep{jiang2023raucg, choi2023large, sarwar2022unsupervised, alexander2023topological}, which can generally be classified into two types: explicit hate speech (ex-HS)~\citep{schmidt2017survey} and implicit hate speech (im-HS)~\citep{elsherief2021latent}. ex-HS refers to straightforward toxic statements, typically featuring derogatory language~\cite{gao-etal-2017-recognizing, waseem2016hateful}. In contrast, im-HS does not contain direct expressions of hate, being a more subtle form to convey prejudice, discrimination, or hatred towards a specific group through sarcasm, insinuation, or other obscured means~\citep{elsherief2021latent, wright2021recast, huang2023chatgpt}.

With the increasing spread of HS on the internet, online platforms have started to control its dissemination~\citep{Twitter2023}, a focus area within content moderation of social bots~\citep{venkatesh2024, park2024}. Due to the explicit nature of ex-HS, detection methods can achieve high detection rates~\citep{lu2023facilitating, roychowdhury2023data, caselli2020feel}. In contrast, im-HS involves sophisticated encoding rules that make it easier to evade automatic detection~\citep{gunturi2023toxvis, wiegand2021implicitly, yin2022hidden}, contributing to its widespread on social media.

There has been a growing body of research actively dedicated to combating the spread of im-HS~\citep{masud2023focal, cao2023prompting, pal2022combating, khan2022white, vargas2021contextual,DBLP:conf/wassa/XiangMYG21}. Existing studies primarily focus on distinguishing im-HS from ex-HS~\citep{kim-etal-2024-label, hartmann2024watching} or other categories that are hard to distinguish, such as offensive and abusive language~\citep{caselli2020feel, wiegand-etal-2022-identifying}, as well as natural language explanations for why an im-HS could be hateful~\citep{yadav-etal-2024-tox, huang2023chain}. Researchers further developed datasets for im-HS in many languages~\citep{sap-etal-2020-social, JIANG2022100182, risch-etal-2021-overview, kim-etal-2024-click, saroj-pal-2020-indian}. 
However, these studies do not identify what makes im-HS implicit and have not validated these patterns across multiple languages in LLMs.

To fill these gaps, we facilitate im-HS detection by explicitly encoding \textit{codetypes} in LLMs. Specifically, codetypes are rhetorical strategies extracted from im-HS that involve the moderation of language and the application of verbal techniques~\citep{jiang2019chinese}. As illustrated in Figure \ref{fig:ideaFramework}, we propose a taxonomy of six codetypes commonly associated with im-HS and use it to enhance LLMs to detect such language. Our experiments on Chinese and English datasets~\citep{lu2023facilitating, elsherief2021latent, ocampo2023depth} show that utilizing codetypes consistently improves im-HS detection rates, highlighting the significance of incorporating knowledge about language dynamics into LLMs. We hope this work and its findings provide more effective tools and theoretical insights for combating im-HS.

\section{Related work}
\label{sec: related work}

\paragraph{Implicit hate speech taxonomy.}
Within existing datasets, HS divisions are generally conducted from two perspectives: 1) the sentiment conveyed by the text, e.g.,~\citet{kulkarni2023revisiting} categorized HS into Hateful, Offensive, Provocative, and Neutral; 2) the target groups, e.g.,~\citet{hartvigsen2022toxigen} subdivided the targets of HS into 13 categories including Black, Mexican, Physically Disabled, LGBTQ+, and others. These taxonomies mainly focus on distinguishing between hate and not hate, with very few studies proposed for im-HS. For im-HS,~\citet{elsherief2021latent} classified sentences into seven groups based on social science and NLP literature: grievance, incitement, inferiority, irony, stereotypical, threatening, and other. However, this taxonomy lacks a unified classification criterion, making it unclear how these categories are related, and difficult to apply to other datasets.

\paragraph{Leveraging external knowledge.}
Some research has explored leveraging external knowledge in HS detection. 
For example,~\citet{clarke2023rule} introduced an exemplar-based contrastive learning approach, using logical rules for content moderation. Nonetheless, this method relies on high-quality rules and examples, resulting in relatively high costs. For im-HS,~\citet{ghosh2023cosyn} incorporated user dialogue context and network features. However, this approach heavily relies on the user's personal history and social background, raising concerns about personal privacy leakage in practical applications.~\citet{lin2022leveraging} utilized contextual background information from Wikipedia\footnote{\url{https://en.wikipedia.org/wiki/Main\_Page}}. While Wikipedia provides summaries for specific entities, many encoding forms in im-HS, such as abbreviations or loanwords~\citep{lu2023facilitating}, lack corresponding definitions, leading to limited performance improvements. Moreover, the lack of validation on datasets across different languages makes it challenging to guarantee the robustness of these methods.

Overall, im-HS detection still faces three challenges: 1) developing a fine-grained taxonomy with clear classification criteria;
2) incorporating external knowledge that can be adaptable to different datasets without requiring excessive computational resources; 
and 3) validating models across datasets in different languages. Our study introduces a novel classification paradigm: codetypes, targeting the specific encoding forms within im-HS. 
By integrating knowledge related to these codetypes, we enhance the model's precision in detecting im-HS across both Chinese and English datasets.

\begin{CJK*}{UTF8}{gbsn}
\section{Codetype strategy}
\label{subsec:Taxonomy of Hate Speech}

We define codetypes as \textit{hate speech encoding strategies} for classifying the type of im-HS an instance belongs to. As there is no unanimous conclusion in socio-linguistics regarding the encoding methods for im-HS due to the diversity of coding objects and language forms, we propose a taxonomy of six critical encoding strategies based on our observations of similarities in the expression of emotions and internal rhetoric across different languages. We find these encoding strategies in both Chinese and English datasets, which implies the existence of cross-linguistic commonalities.

\begin{figure}[t]
\centering
  \includegraphics[width=\linewidth]{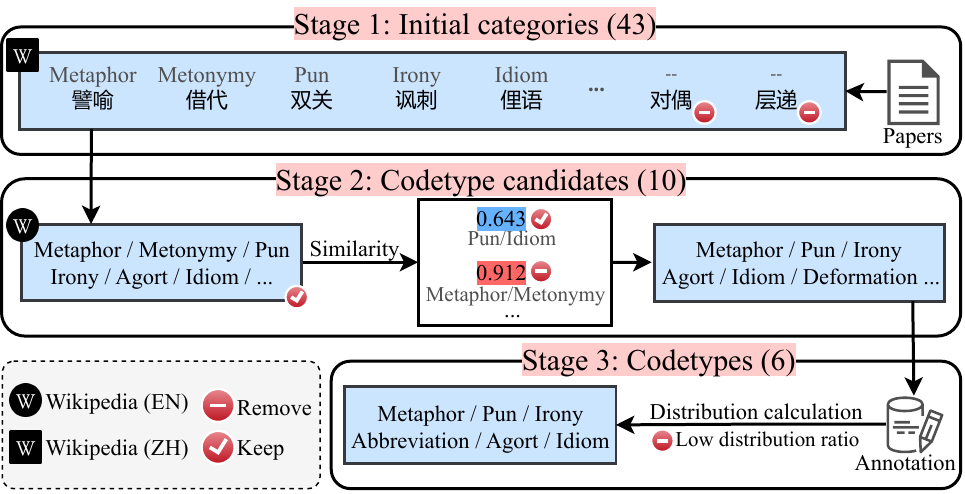}
  \caption{The pipeline for codetype taxonomy construction, with the blue boxes at each stage representing the categories filtered through the selection process.}
   \label{fig:pipeline}
\end{figure}

\subsection{Taxonomy methodology}
\label{sec: Taxonomy}

By our definition, codetypes are rhetorical strategies extracted from im-HS that involve the moderation of language and the application of verbal techniques~\citep{jiang2019chinese}. To delineate the taxonomy of im-HS, we employ a systematic process to construct the suitable codetype taxonomy that underlay im-HS detection. Since existing studies on rhetorical strategy classification already provide a comprehensive and structured system for Chinese corpora~\citep{lu1993study,lu2004rhetoric,kirkpatrick2012chinese}, we first select an initial list of codetype candidates (in Chinese) from the~\textit{rhetorical styles}~\citep{jiang2019chinese} and the~\textit{formation modes of new internet words}~\citep{jing2019chinese,tao2017investigation}; candidates not found in Chinese Wikipedia are filtered out, leaving 43 codetype categories. Then, we filter these categories using English Wikipedia, removing codetypes that lack a corresponding name or explanation in English. In the next step, we encode the remained codetypes with their explanations in Chinese Wikipedia using a pre-trained word2vec model~\citep{mikolov2013efficient}; we calculate cosine similarity among all codetypes using their word2vec embeddings and eliminate those with a similarity score higher than 0.9, down-sampling the list of codetype candidates to 10 categories: \textit{Irony}, \textit{Metaphor}, \textit{Argot}, \textit{Pun}, \textit{Abbreviation}, \textit{Idiom}, \textit{Rhetorical question}, \textit{Loanword}, \textit{Hyperbole}, and \textit{Deformation}.
The similarities between these codetype candidates are shown in Figure~\ref{fig:heatmap} in Appendix~\ref{app: Annotation distribution}.
Additionally, we include a \textit{None} category for instances that do not fall into any of the previous categories.

To ensure the quality of the proposed taxonomy, we conduct a manual verification on 200 samples randomly selected from ToxiCN~\citep{lu2023facilitating} (more details in Section~\ref{sec:Experiments}). Subsequently, three annotators are hired to classify these samples into the 11 categories mentioned above, with inter-annotator agreement Fleiss' kappa~\citep{fleiss1971measuring}, $\kappa = 0.43$ (moderate agreement). We introduce a fourth annotator to resolve disagreement on difficult cases. The detailed annotation guidelines can be found in Appendix~\ref{app: annotation guidelines}. The distribution of the categories in the final annotated subset is shown in Figure~\ref{fig: annotation}, with a consensus ratio over 75\%, as detailed in Table~\ref{tab:annotation} in Appendix~\ref{app: Annotation distribution}. The top six most frequent categories account for approximately 80\% of the samples, whereas \textit{rhetorical question}, \textit{loanword},  \textit{hyperbole}, and \textit{deformation} together are all less than 5\%. Based on these results, we construct the final codetype taxonomy with the top six most frequent categories:~\textit{Irony}, \textit{Metaphor}, \textit{Pun}, \textit{Argot}, \textit{Abbreviation}, and \textit{Idiom}. Additionally, we include an \textit{Other} category for cases that do not fit into the aforementioned codetypes.

\begin{figure}[t]
  \includegraphics[width=\columnwidth]{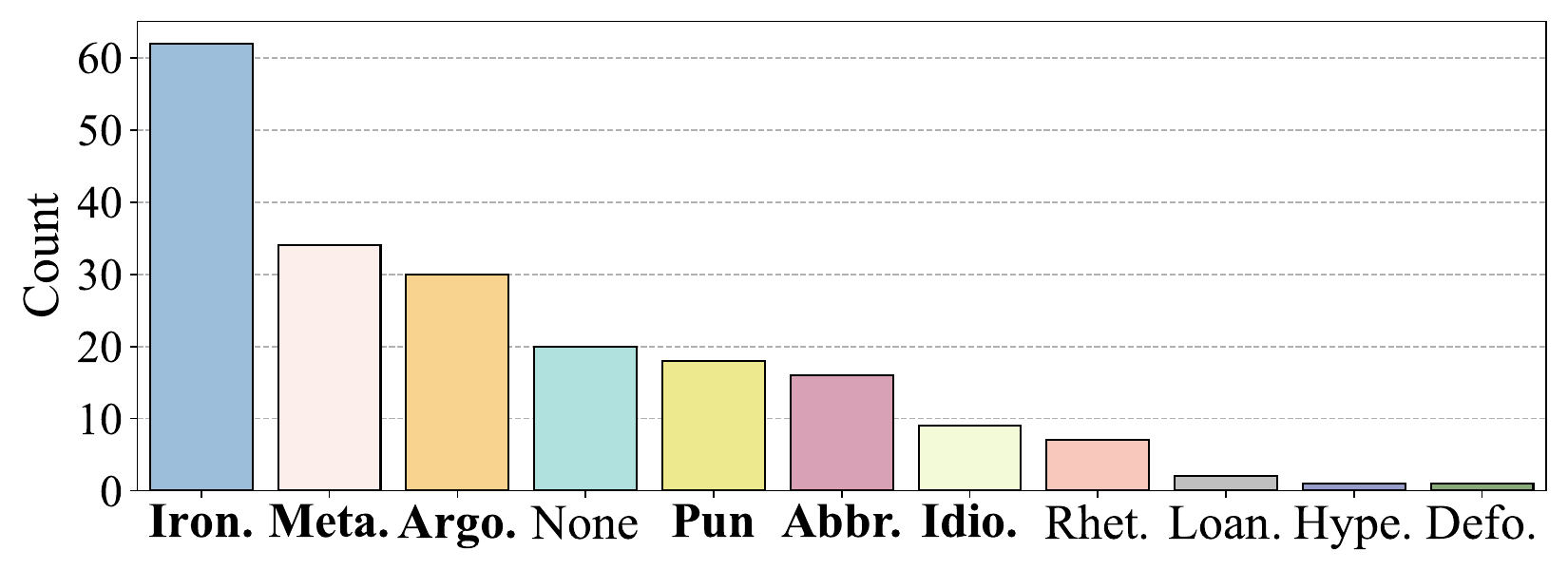}
  \caption{Distribution of codetype candidate categories in a subset of ToxiCN dataset. The six codetypes that are used in the final taxonomy are highlighted in \textbf{bold}.}
    \label{fig: annotation}
\end{figure}

\begin{table*}[h]
\centering
\begin{adjustbox}{width=\linewidth}
\begin{tabular}{llll}
\toprule
\textbf{Dataset} & \textbf{Name} & \textbf{Explanation in Wikipedia} & \textbf{Sample}\\
\toprule
ToxiCN  & 缩写 & \makecell[l]{缩写是在使用拼音或文字的语言中，对于常\\用的词组以及少数常用的词所采用的简便写法。\\ \textit{Abbreviation is a simplified way of writing commonly} \\ \textit{used phrases and a few frequently used words in lan-} \\ \textit{guages that use pinyin or characters.}}
  & \makecell[l]{ \underline{txl}是社会问题，应该有人管，而不是当透明人 \\ \textit{\underline{\pinyin{tong2xing4lian4}} (Homosexual) is a social issue and someone} \\ \textit{should address it, rather than being invisible. } } \\

\bottomrule

Latent /\ ISHate & Abbreviation & \makecell[l]{Abbreviation is a shortened form of a word or phrase,\\ by any method.} & \makecell[l]{\underline{WPWW}} \\
\bottomrule
\end{tabular}
\end{adjustbox}
\caption{Examples for the abbreviation in different datasets. The keywords related to abbreviations within the samples are \underline{underlined}. English translations for ToxiCN dataset are shown below for reference. Specifically, \underline{txl} is an abbreviation derived from its corresponding pinyin, while \underline{WPWW} stands for White Pride World Wide.}
\label{tab:example}
\end{table*}

\subsection{Codetype definitions}
The definitions for each codetype in our proposed taxonomy are:

\begin{description}

\item[Abbreviation] pertains to a shortened form of a word or phrase, and often constitutes a convenient form of writing for commonly used phrases (mostly proper nouns) and a few frequently used words. 
For example in English, \textit{kkk} is used to represent the Ku Klux Klan, an extremely racist and white supremacist group. Examples can be found in Table~\ref{tab:example}.

\item[Metaphor] is a rhetorical strategy that connects unrelated concepts to create novel associations. Speakers often use it by comparing the target group to a distinct group of objects, such as animals, or connecting the target group with behaviors that are not commonly accepted, such as animalistic behaviors, or tangible events as descriptive analogies. It encompasses both direct and indirect manifestations. 
For example, the phrase \textit{大肥猪} (\textit{big fat pig}) is used in Chinese online communities to mock overweight women. Similar patterns have also been observed in western online community, e.g., using~\textit{big whale} to fat-shaming women.

\item[Irony] pertains to an inconsistency between surface and intended meanings, implying a divergence between explicit and implicit messages. Speakers frequently use praiseworthy language ironically, which focuses specifically on instances where the speakers' intended message contrasts with the literal interpretation of the words used. For instance, the phrase \textit{小仙女} (\textit{little fairy}), originally used to describe beautiful women, is now widely used as a misogynistic phrase in current Chinese online community. 

\item[Pun] is a linguistic usage that exploits homophony or analogy, enabling a sentence to indirectly convey alternative meanings. In current online community, users often use homophonic characters to replace certain sensitive words to avoid automatic hate speech detection. For instance, the innocuous Chinese phrase~{亩篝} (pronounced~\pinyin{mu3gou1}) sounds similar to~\textit{母狗} (pronounced~\pinyin{mu3gou3}), which means~\textit{bitch}.

\item[Idiom] is an informal and colloquial phrase utilized in everyday communication, also known as dialects or vernacular language, which often originate from dialects in certain regions and become widely used through continuous usage.
For example, the phrase \textit{基佬} (\textit{gay}), which might be considered as disrespectful, was originally popularized in Hong Kong to refer to homosexual men. 

\item[Argot] refers to language specific to a particular domain or culture, often incomprehensible to those who are not familiar with the background. Online communities often foster their unique expressions and phrases; for instance, the term~\textit{被绿}~(\textit{getting greened}) typically refers to being cheated in a relationship, and only becomes trending in recent years.

\end{description}

It is worth noting that, our proposed codetype taxonomy is not a direct indicator of im-HS, but serves as a structured taxonomy that can help LLMs better understand the context and the actual meaning of the sentences.

\begin{figure}[t]
\centering
    \scriptsize
    \includegraphics[width=\linewidth]{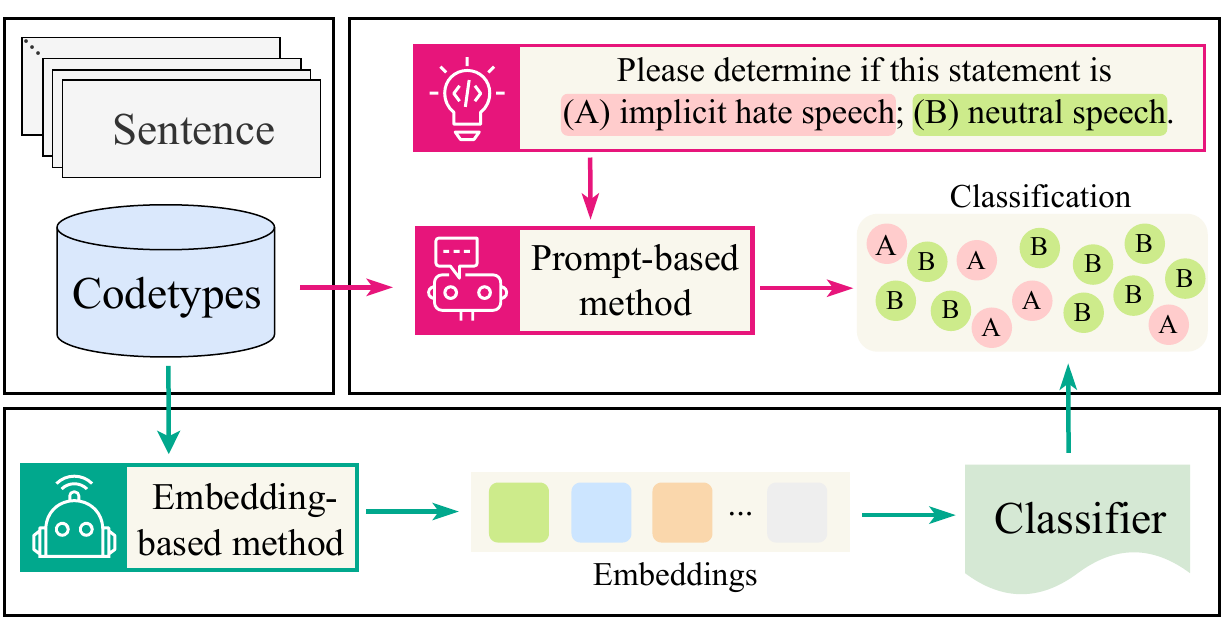}
    \caption{The overall framework of our prompt-based method and embedding-based method.}
    \label{fig: model framework}
\end{figure}

\section{Implicit hate speech detection with codetypes}
\label{sec:Hate Speech Detection with Codetype}
We show the effectiveness of the proposed codetype taxonomy with LLMs for im-HS detection in two different ways, as shown in Figure~\ref{fig: model framework}:

\begin{enumerate}
    \item \textbf{Prompt-based method} employs LLMs directly with prompts to classify text using the generated outputs from models. 
    \item \textbf{Embedding-based method} uses LLMs as frozen encoders, by extracting the hidden states from the inner layers of the models during encoding as features; these features are then fed into a classifier (logistic regression in our case) for im-HS detection.
\end{enumerate}

\subsection{Prompt-based method}
\label{sec:prompt-based method}
Using codetypes as part of the prompts is the most common way to exploit the power of LLMs. Given the $i$-th sample $s_i$ from a dataset $\mathcal{D}$ and $K$ codetypes $C = \{c_1,\cdots, c_K\}$, we leverage an LLM $M$ with instruction $I$ to generate predicted label $l_i$ for im-HS detection. We denote the process of generating $l_i$ using $M$ with $C$ as $f_{\text{prompt}}$ such that:

\begin{equation}
\centering
l_i = f_{\text{prompt}} \left( \left[C;s_i;I\right], M \right)
\label{eq:f-prompt}
\end{equation}

\noindent Since $I$ and $M$ appears universally when using $f_{\text{prompt}}(\cdot)$, and $C$ is the variable of our interest, for simplicity, we omit $I$ as well as $M$ in $f_{\text{prompt}}(\cdot)$. The instruction $I$ is:

\begin{prompt}[title={\textbf{User prompt}}]
Please determine if \texttt{[s]} is (A) implicit hate speech or (B) neutral speech.
\end{prompt}

\noindent If codetypes are included, then they are concatenated with $I$ as prefix:

\begin{prompt}[title={\textbf{User prompt with codetypes}}]
Codetypes are rhetorical strategies extracted from implicit hate speech that involve the moderation of language and the application of verbal techniques. Please answer based on the information of these 6 codetypes:\textcolor{gray}{\texttt{\textbackslash n}}\\
\texttt{[C]}
\textcolor{gray}{\texttt{\textbackslash n}}\\
Please determine if \texttt{[s]} is (A) implicit hate speech or (B) neutral speech.
\end{prompt}

\noindent Here \texttt{[C]} denotes the codetype information, and \texttt{[s]} denotes the sample. More details can be found in the Appendix~\ref{app:prompt-based methd}.

\subsection{Embedding-based method}
While the predominant usages of LLMs are for generative tasks, previous work~\citep{burns2022discovering} has shown that leveraging the information within the hidden layers of LLMs can further improve their performance on downstream tasks. Inspired by these, we leverage the generative LLMs as frozen encoders and use the hidden states of their inner layers as features. Specifically, we follow~\citet{li2024inference} to utilize the output of the multi-head attention (MHA) as features; we use the MHA output of all transformer layers within a model to fully exploit the model. The MHA output from different layers is then concatenated and fed into a trainable classifier.

For a model $M$, its ability of transforming a piece of text $s_i$ into a corresponding embedding $\mathbf{E}_{s_i}$ can be expressed as a function $f_{\text{emb}}(\cdot)$:

\begin{equation*}
\mathbf{E}_{s_i} = f_{\text{emb}}(s_i)
\end{equation*}

\noindent Similar to Equation~\ref{eq:f-prompt}, we omit $M$ for simplicity. Considering that $s_i$ can potentially encompass multiple codetypes, we integrate all available candidates from $C$ with $s_i$ to serve as input for $M$. Specifically, we design three strategies for embedding construction to ensure that all codetype information is properly encoded, as depicted in Figure~\ref{fig: embedding framework}. Now we explain each of them in details:

\begin{figure}[t]
\centering
\vspace{-5.5mm}
  \includegraphics[width=5.4cm]{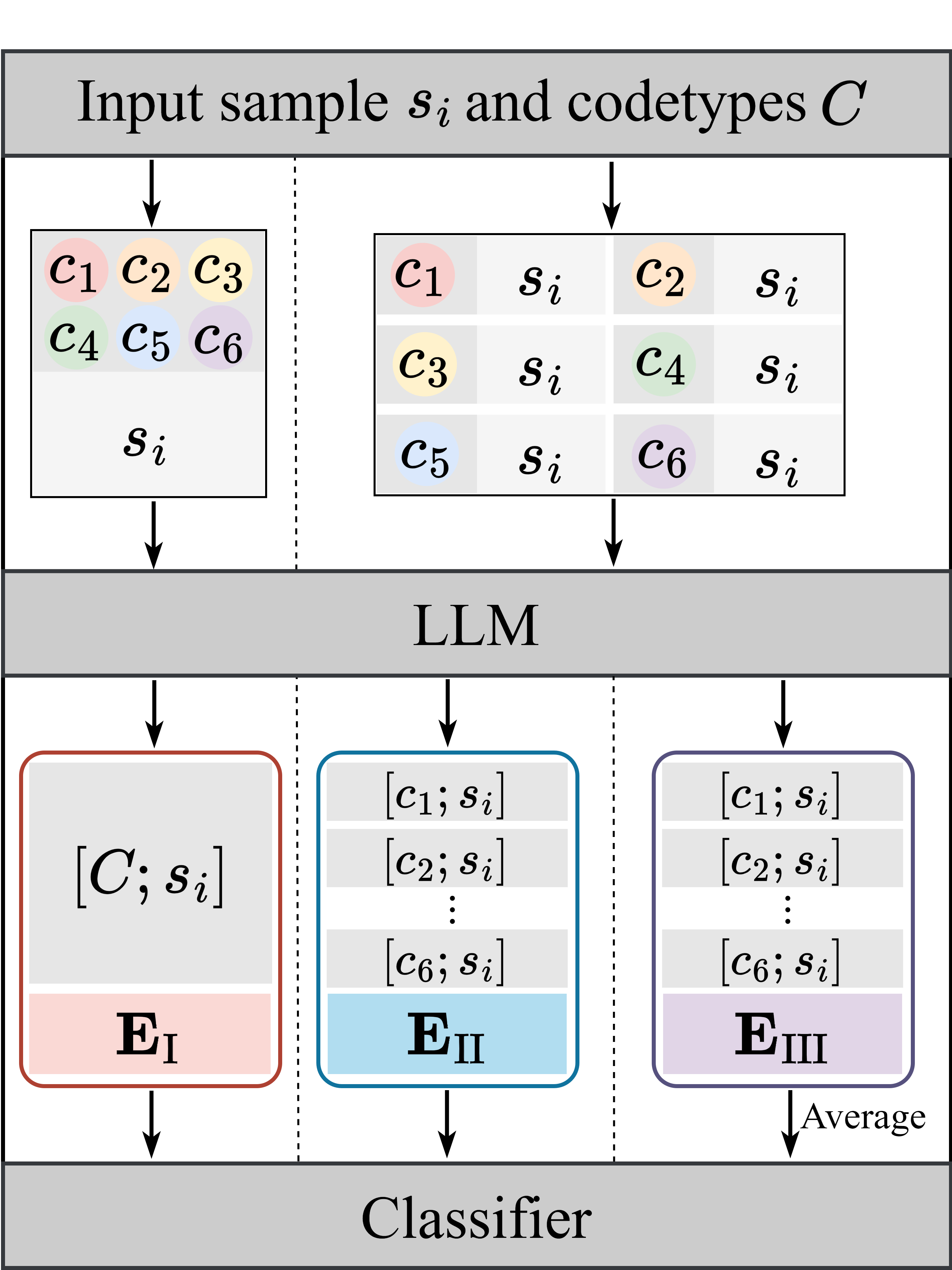}
  \caption{The overall framework of the three proposed embedding methods. Here $c_1, \dots, c_K$ represent the $K$ codetypes ($K=6$ in our case), and $S$ denotes the sentence to be classified.}
    \label{fig: embedding framework}
\end{figure}

\paragraph{Method I.}\quad We directly concatenate the codetypes $C$ with the sample $s_i$ as input and then feed it into the model $M$:
\vspace{-0.5cm} 

\begin{equation*}
\mathbf{E}_{s_i} = f_{\text{emb}}(\left[ C; s_i \right])
\end{equation*}

\paragraph{Method II.}\quad
We first combine each codetype $c_k \in C \ ( 1 \leq k \leq K )$ with $s_i$ individually, then feed each of them into $M$ to get an embedding $\mathbf{E}_{s_i}^k$, and concatenate all of them to get the final embedding:

\begin{equation*}
\begin{split}
\mathbf{E}_{s_i} &= \left[f_{\text{emb}}(\left[ c_1; s_i \right]), \cdots, f_{\text{emb}}(\left[ c_K; s_i \right]) \right]\\
&=\left[\mathbf{E}_{s_i}^1, \cdots, \mathbf{E}_{s_i}^K \right]\\
\end{split}
\end{equation*}

\paragraph{Method III.}\quad
Similar to Method 2, Method 3 also combine each codetype with $s_i$ individually first; but instead of applying concatenation to the embeddings, this method applies element-wise average over all obtained embeddings to get the final embedding:
$$
\mathbf{E}_{s_i} = \frac{1}{K} \sum_{k=1}^{K} \mathbf{E}_{s_i}^k
$$

In the following sections, we denote the embedding produced via these three methods as $\mathbf{E}_{\text{I}}$, $\mathbf{E}_{\text{II}}$, and $\mathbf{E}_{\text{III}}$ correspondingly.

\begin{table}[t]
\centering
\begin{adjustbox}{width=\linewidth}
\begin{tabular}{rrrrr}
\toprule
\textbf{Dataset} & \textbf{Lang.} & \textbf{Im-hate} & \textbf{No-hate} & \textbf{Total}\\
\toprule
ToxiCN & ZH & 5,645 & 5,550 & 11,195\\
Latent & EN & 7,100 & 13,291 & 20,391 \\
ISHate & EN & 1,238 & 17,869 & 19,107\\
\bottomrule
\end{tabular}
\end{adjustbox}
\caption{Statistics on implicit hate speech (im-hate) and no-hate speech for different datasets. Here \textbf{Lang.} stands for languages of the datasets.}
\label{tab:statistic}
\end{table}

\section{Experiments}
\label{sec:Experiments}
We first introduce the datasets and models, and then evaluate the effectiveness of our proposed codetypes on the im-HS detection task.
\subsection{Datasets}
\label{subsec:Datasets}

We select three datasets in two languages for our experiments: ToxiCN~\citep{lu2023facilitating} in Chinese, while the Latent-hatred~\citep{elsherief2021latent} and ISHate~\citep{ocampo2023depth} dataset in English. Details for each dataset are presented in Table~\ref{tab:statistic}.

\paragraph{ToxiCN}\quad ToxiCN is derived from posts published on two Chinese public online platforms: Zhihu\footnote{\url{https://www.zhihu.com/}} and Tieba\footnote{\url{https://tieba.baidu.com/index.html}}, which cover sensitive topics such as gender, race, regional issues, and LGBTQ+. 

\paragraph{Latent-hatred}\quad Latent-hatred is a commonly used dataset for English im-HS detection tasks. It consists of tweets shared by online hate groups and their followers on Twitter.

\paragraph{ISHate}\quad ISHate builds on the seven English hate speech datasets and for the first time provides a more nuanced categorization for HS, including both implicit and subtle ones. 

\subsection{Models}

We use open-sourced models that are trained on both Chinese and English for our experiments:

\paragraph{Baichuan2-13B-Chat}\quad Baichuan2-13B-Chat~\citep{Baichuan2-13B-Chat} is a 13B LLM trained on a corpus with 2.6 trillion tokens and is reported to have achieved the best performance in several Chinese and English benchmarks.

\paragraph{Llama2-Chinese-Chat}\quad Llama2-Chinese-Chat~\citep{Llama2-Chinese} is a series of models developed based on Llama2 models from~\citet{DBLP:journals/corr/abs-2307-09288}, which are then further fine-tuned using Chinese instruction-following datasets. Specifically, we use the 7B and 13B checkpoints in our experiments.

Logistic regression model is used as the classifier for the embedding-based method. Notice that for prompt-based method, all models are not frozen; for embedding-based method, only the logistic regression classifier is trained and the LLMs are frozen with no parameter updates.

\begin{table*}[ht]
\centering
\small
\begin{adjustbox}{width=\textwidth}
\begin{tabular}{c|l|ccc|ccc|ccc}
\toprule
\multicolumn{1}{c|}{\raisebox{-0.8\normalbaselineskip}[10pt][10pt]{\begin{tabular}{c} \textbf{Method} \\ \end{tabular}}} 
& \multicolumn{1}{c|}{\raisebox{-0.8\normalbaselineskip}[0pt][0pt]{\begin{tabular}{c} \textbf{Codetype} \\ \end{tabular}}} 
& \multicolumn{3}{c|}{\textbf{ToxiCN}} 
& \multicolumn{3}{c|}{\textbf{Latent-hatred}} 
& \multicolumn{3}{c}{\textbf{ISHate}}\\
\cline{3-11}

\multicolumn{1}{c|}{} & & \textbf{Bai2-13B} & \textbf{Llama2-7B} & \textbf{Llama2-13B} & \textbf{Bai2-13B} & \textbf{Llama2-7B} & \textbf{Llama2-13B} & \textbf{Bai2-13B} & \textbf{Llama2-7B} & \textbf{Llama2-13B} \\
\hline

{\raisebox{-3.3\normalbaselineskip}[0pt][0pt]{Prompt}} 
& \textbf{-} &0.2556&0.5950&0.6634&0.3872&0.0988&0.3331&0.1188&0.1805&0.1189\\ 
& {Name} &0.2219&0.4739&0.6494&0.3463&0.1935&0.3352&0.1101&0.1404&0.0926\\ 
& {Expl} &0.1109&0.5182&0.6454&0.3822&0.2400&0.4070&0.1123&0.0808&0.1214\\ 
& {Samp} &0.2988&0.4426&0.6503&0.3964&0.3750&0.3641&0.1030&0.0973&0.1116\\ 
& {Name+Expl} &0.1923&0.5657&0.6307&0.3519&0.2857&0.3161&0.1077&0.1235&0.1012\\ 
& {Name+Samp} &0.2763&0.5407&0.6396&0.3995&0.2222&0.3366&0.1038&0.1538&0.1151\\ 
& {Samp+Expl} &0.1905&0.5538&0.6203&0.3929&0.2857&0.4318&0.1211&0.3333&0.1147\\ 
& {Name+Samp+Expl} &0.1918&0.5931&0.6225&0.4232&0.1700&0.4055&0.1122&0.1875&0.1274\\ 
\hline

$\mathbf{E}_{no}$ & - &0.7405&0.7593&0.7679&0.5798&0.6282&0.6275&0.5505&0.6625&0.6567\\
\hline

{\raisebox{-3.3\normalbaselineskip}[0pt][0pt]{$\mathbf{E}_{\text{I}}$}} 
& {Name} &0.6994&0.7893&0.7663&0.5092&0.7010&0.6834&0.3911&0.6990&0.6709\\ 
& {Expl} &0.6768&0.7766&0.7638&0.4508&\textbf{0.7034}&\underline{0.6867}&0.2918&0.6667&0.6625\\ 
& {Samp} &0.7037&0.7890&\underline{0.8054}&0.4799&0.6939&0.5867&0.4069&0.6583&0.6383\\ 
& {Name+Expl} &0.7012&0.7431&0.7930&0.4470&0.5733&0.6708&0.2468&0.6749&0.6522\\ 
& {Name+Samp} &0.6799&0.7854&0.7713&0.4877&0.6490&0.6225&0.2869&0.6892&0.6498\\ 
& {Samp+Expl} &0.6661&\textbf
{0.7982}&0.7945&0.4523&0.6225&0.5949&0.1659&0.6688&0.6506\\ 
& {Name+Samp+Expl} &0.6684&0.7623&\textbf{0.8091}&0.4272&0.6395&0.6076&0.1435&0.6892&0.6522\\ 
\hline

{\raisebox{-3.3\normalbaselineskip}[0pt][0pt]{$\mathbf{E}_{\text{II}}$}} 
& {Name} &0.7550&0.7783&0.7710&0.5401&0.6795&0.6608&\textbf{0.5894}&0.6923&\textbf{0.7055}\\ 
& {Expl} &0.7322&0.7804&0.7536&0.5479&0.6672&0.6584&0.5342&0.7087&0.6748\\ 
& {Samp} &0.7161&\underline{0.7907}&0.7727&0.5554&0.5867&0.6623&0.5519&0.6972&0.6967\\ 
& {Name+Expl} &0.7380&0.7656&0.7733&0.5525&0.6585&0.6415&\underline{0.5878}&0.6988&\underline{0.6988}\\ 
& {Name+Samp} &0.7436&0.7810&0.7907&0.5416&0.6275&0.6351&0.5220&0.7112&0.6888\\ 
& {Samp+Expl} &0.7183&0.7850&0.7857&0.5529&0.6887&0.6839&0.4762&0.7139&0.6728\\ 
& {Name+Samp+Expl} &0.7329&0.7512&0.7838&0.5294&0.6410&0.6711&0.4797&0.7112&\underline{0.6988}\\
\hline
 
{\raisebox{-3.3\normalbaselineskip}[0pt][0pt]{$\mathbf{E}_{\text{III}}$}} 
& {Name} &\underline{0.7687}&0.7854&0.7821&\textbf{0.6073}&0.6998&\textbf{0.6882}&0.5831&0.7055&0.6563\\ 
& {Expl} &0.7475&0.7824&0.7888&0.5560&\underline{0.7029}&0.6807&0.5694&\textbf{0.7305}&0.6707\\ 
& {Samp} &0.7446&0.7870&0.7822&\underline{0.5886}&0.6225&0.6494&0.5526&0.6829&0.6890\\ 
& {Name+Expl} &\textbf{0.7786}&0.7736&0.7658&0.5679&0.6621&0.6667&0.5180&0.6848&0.6768\\ 
& {Name+Samp} &0.7436&0.7560&0.7650&0.5630&0.6623&0.6447&0.5552&\underline{0.7156}&0.6667\\ 
& {Samp+Expl} &0.7324&0.7547&0.7945&0.5451&0.6241&0.6835&0.5000&0.7130&0.6729\\ 
& {Name+Samp+Expl} &0.7520&0.7570&0.7822&0.5647&0.6395&0.6709&0.4855&\underline{0.7156}&0.6768\\ 
\bottomrule
\end{tabular}
\end{adjustbox}
\caption{\label{tab:exp results} Experiment results evaluated using F1 score. $\mathbf{E}_{no}$ represent the embedding-based method without adding codetypes. The \textbf{codetype} column shows the combination of three types of codetype information. The best results for each model are highlighted in~\textbf{bold}, while the second best results are \underline{underlined}.}
\end{table*}

\subsection{Experimental setup}
We divide each dataset into training, validation, and testing sets with a ratio of 8:1:1. We set the learning rate to 5e-4 and choose Adam~\citep{kingma2014adam} as the optimizer. We select F1 score as the evaluation metric for measuring im-HS detection results as the datasets are usually not balanced between categories. To exploit the best way of utilizing codetype, we categorize codetype-related information into three components:
\begin{enumerate}
    \item \textbf{Name}: the name of the codetype.
    \item \textbf{Expl}: codetype explanation on Wikipedia.
    \item \textbf{Samp}: selected sample for the corresponding codetype.
\end{enumerate}
\noindent Examples of codetype-related information are shown in Table~\ref{tab:example}. For each input sentence, the codetype information is preassigned, consisting of six codetypes. We explore different combinations of these codetype information on both prompt-based method and embedding-based method. We apply no codetype information in the baseline.

\subsection{Classification results}

The results in Table~\ref{tab:exp results} indicate an enhancement in classification performance across three datasets with the inclusion of codetype information.

\paragraph{Prompt-based method vs. embedding-based method}\quad There exists a huge performance gap between the prompt-based method and embedding-based methods when using the same model, particularly on the ISHate dataset. This difference becomes the most significant when using Llama2-Chinese-13B-Chat, where the gap between the prompt-based method and the $\mathbf{E}_{\text{II}}$ method reaches an average F1 score of about 58\%. The average F1 score of the prompt-based method is generally lower in ToxiCN comparing to its baseline; however, in the two English datasets Latent-hatred and ISHate, prompt-based method with some combinations of codetype information do present superior performance in comparison with baseline.
On the contrary, the embedding-based methods show improvements over baseline after the introduction of codetypes. The highest average improvement can be observed in the $\mathbf{E}_{\text{I}}$ method, where the Llama2-Chinese-7B-Chat model achieves approximately a 7.5\% F1 score increase on the Latent-hatred dataset compared to its baseline. Meanwhile, Llama2-Chinese-13B-Chat achieves the best performance of 0.8091 F1 score using Name+Samp+Expl combination and $\mathbf{E}_{\text{I}}$ method. For different embedding-based methods, we find that the average F1 score improvement against baseline of $\mathbf{E}_{\text{II}}$ and $\mathbf{E}_{\text{III}}$ are higher than $\mathbf{E}_{\text{I}}$, by 5.63\% and 5.09\%, respectively. This indicates that $\mathbf{E}_{\text{II}}$ and $\mathbf{E}_{\text{III}}$ own better robustness than other methods.

\begin{table*}[h]
\centering
\begin{adjustbox}{width=\linewidth}
\begin{tabular}{lllll}
\toprule
\textbf{Statement} & \textbf{Codetype} & \textbf{Label} & \textbf{Predict} & \textbf{Dataset}\\
\toprule
\makecell[l]{(a) why not take a look at haiti or liberia or somalia or chad where black people \\ were free from redlining ?}  &  None &  no-hate  & im-hate & Latent-hatred \\
\bottomrule

\makecell[l]{(b) \underline{xswl}楼主这签名 \\ \textit{ \underline{LOL}, the OP's signature.}} & Abbreviation, Argot & no-hate & im-hate & ToxiCN \\
\bottomrule

\makecell[l]{(c) Why did I get a \underline{white car}? So it would actually work!} & Pun, Metaphor & im-hate & no-hate & ISHate \\
\bottomrule

\makecell[l]{(d) \underline{图沙黑蛆}是人类未来大势所趋 \\ \textit{ \underline{The massacre of black maggots} is the inevitable trend for humanity's future.}} & Pun, Metaphor  & im-hate & no-hate & ToxiCN \\
\bottomrule

\end{tabular}
\end{adjustbox}
\caption{Example statements of incorrect predictions from different datasets. English translations for ToxiCN examples are shown below for reference. The keywords related to codetypes within the statements are \underline{underlined}.}
\label{tab: wrong example}
\end{table*}

\paragraph{Effectiveness of codetype combinations in the prompt-based method} 
When comparing the performance within methods, we count the number of best F1 scores across different codetype combinations. Specifically, the prompt-based method tends to score the highest in combinations that include samples, including Samp (33.3\%), Samp+Expl (33.3\%), and Name+Samp+Expl (33.3\%). Introducing samples in the prompt-based method acts as a few-shot learning approach for LLMs. However, most of the best scores are not achieved with Samp alone but rather when combined with the codetypes Name and Expl, suggesting that the prompt incorporating codetype names and explanations provides LLMs with more learning rooms, thereby enhancing classification. 

\paragraph{Performance comparison of embedding-based methods across Chinese and English datasets} Additionally, we find that the average F1 scores for the embedding-based methods are higher on the Chinese dataset (ToxiCN: 0.7598) than on the English datasets (Latent-hatred: 0.6121, ISHate: 0.6062). We also observe that Llama2-Chinese-7B-Chat and Llama2-Chinese-13B-Chat generally outperform the Baichuan2-13B-Chat model across different datasets. This difference is pronounced in the English datasets. For instance, the best scores of Llama2-Chinese-7B-Chat on the Latent-hatred and ISHate datasets exceed the best scores of Baichuan2-13B-Chat within $\mathbf{E}_{\text{I}}$ method by 19.42\% and 29.21\%, respectively.

\subsection{Result analysis}
\paragraph{$\mathbf{E}_{\text{II}}$ and $\mathbf{E}_{\text{III}}$ are superior to $\mathbf{E}_{\text{I}}$}\quad Among the three embedding-based methods, $\mathbf{E}_{\text{I}}$ concatenates all codetype information with the statements at once, whereas $\mathbf{E}_{\text{II}}$ and $\mathbf{E}_{\text{III}}$ combine each codetype with the statement individually. This allows the model to better match and verify each codetype with the statement.

\paragraph{LLMs perform worse with Name+Samp+Expl combinations compared to using Name or Samp}\quad Although the introduction of external information can enhance the model's classification performance, it often leads to overcorrection issues~\citep{lin2022leveraging, lu2023facilitating}. This explains why the LLMs perform better when only introducing single combinations like Name or Samp, compared to combinations such as Name+Expl, Name+Samp, Samp+Expl, or Name+Samp+Expl. When provided with more codetype information beyond Name or Samp, the model is more likely to misclassify neutral statements related to sensitive groups as implicit hate, e.g., statement (a) in Table~\ref{tab: wrong example}, or misinterpret the codetype information in the statement, leading to the misclassification of neutral statements containing codetypes as implicit hate, e.g., statement (b) in Table~\ref{tab: wrong example}.

\paragraph{More codetypes lead to decreased LLM Performance}
A statement often contains more than one codetype, e.g., statements (b)-(d) in Table~\ref{tab: wrong example}. Specifically, \textit{xswl} in statement (b) is a Chinese pinyin abbreviation of \textit{LOL} and is widely used in Chinese social media. In statement (c), \textit{white car} is a metaphor for white supremacy. This statement may appear as a harmless joke on the surface, but it can also carry implicit hate if interpreted in a different context, with \textit{work} potentially drawing on racial undertones about whiteness and superiority. Additionally, in statement (c), the Chinese pronunciation of \textit{图沙} is similar to \textit{massacre}, and \textit{black maggots} is a metaphor for black people. We find that as the number of codetypes increases, the difficulty of accurately interpreting statements rises.

\section{Conlusions}
Our research introduces a novel strategy for detecting im-HS, proposing a codetype taxonomy that encompasses various strategies encoding implicit hateful intentions. We develop a systematic process to finalize the six codetype categories.
Additionally, to validate the performance improvements brought by the introduction of codetypes, we propose two methods: prompt-based method and embedding-based method. The two methods are tested on three models trained on both Chinese and English corpora, using different combinations of codetypes to evaluate the effectiveness. In our experimental result analysis, we compare the performance improvements between the two methods and further analyze the strengths and weaknesses of different models, codetype combinations, and embedding-based methods. Experimental results from both Chinese and English datasets establish the efficacy of incorporating codetype information into LLMs, enhancing the effectiveness of im-HS detection.

\section*{Limitations}
The comprehensiveness of our codetype taxonomy and the applicability of our methods across linguistic contexts remain areas for further exploration. Furthermore, we recognize that the selection of codetype samples also influences experimental performances and our methodologies have limited capability in handling more complex im-HS. Building a model that can dynamically determine the appropriate codetypes based on the input sentence would enhance the efficiency of the detection process. For statements containing more than two codetypes, further optimization of the model is required. For instance, using a chain-of-thought approach to prompt the LLM can help improve its classification performance on the prompt-based method.

\section*{Ethical considerations}
While we prioritize the efficacy of im-HS detection leveraging codetype information, we acknowledge the critical importance of addressing ethical considerations within our research. Despite our efforts to provide warnings regarding potential instances of offensive or vulgar content, the presentation of implicit hate examples may inadvertently cause psychological distress to readers. Furthermore, there is a risk that these examples could be exploited by LLMs, thereby contributing to harmful discourse on a broader scale.

It is essential to clarify that our research aims to enhance the classification accuracy of LLMs for im-HS. While combating the proliferation of hate speech requires continuous effort, our exploration on linguistic patterns within im-HS both deepens our understanding of the phenomenon and demonstrates the potential for improved detection across diverse language datasets.

\section*{Acknowledgments}
This work was partly supported by JSPS KAKENHI No. JP22K12091.

\bibliography{custom}

\appendix

\newpage
\section{Annotation guidelines}\label{app: annotation guidelines}

We employ four graduate and undergraduate students majoring in computer science and statistics as annotators. Their primary responsibility is to categorize $10$ Chinese codetype candidates and determine the definitive codetypes. The annotated data is randomly selected from the ToxiCN dataset, which encompasses a total of $200$ posts from users on Zhihu and Tieba.

\subsection{Distribution of 10 codetype candidates}\label{app: Annotation distribution}

\begin{table*}[htb]
\centering
\begin{adjustbox}{width=\textwidth}
\begin{tabular}{c|c|c|c|c|c|c|c|c|c|c|c}
\toprule
\multicolumn{1}{c|}{} & \textit{Irony} & \textit{Metaphor} & \textit{Argot} & \textit{Pun} & \textit{Abbreviation} & \textit{Idiom} & \textit{Rhetorical*} & \textit{Loanword} & \textit{Hyperbole} & \textit{Deformation} & \textit{None} \\
\hline
\multicolumn{1}{c|}{Annotator 1}
& 49 & 26 & 30 & 16 & 10 & 6 & 29 & 4 & 18 & 1 & 11 \\
\hline
\multicolumn{1}{c|}{Annotator 2}
& 63 & 31 & 26 & 20 & 16 & 10 & 13 & 2 & 7 & 1 & 11 \\
\hline
\multicolumn{1}{c|}{Annotator 3}
& 78 & 24 & 22 & 18 & 17 & 7 & 15 & 2 & 5 & 1 & 11 \\
\hline
\multicolumn{1}{c|}{Consensus}
& 54 & 26 & 17 & 13 & 16 & 8 & 7 & 2 & 1 & 1 & 8 \\
\hline
\multicolumn{1}{c|}{Annotator 4}
& 8 & 8 & 13 & 5 & - & 1 & - & - & - & - & 12 \\
\hline
\multicolumn{1}{c|}{\textbf{Final}} & \textbf{62} & \textbf{34} & \textbf{30} & \textbf{18} & \textbf{16} & \textbf{9} & \textbf{7} & \textbf{2} & \textbf{1} & \textbf{1} & \textbf{20} \\
\bottomrule
\end{tabular}
\end{adjustbox}
\caption{\label{tab:annotation}Statistics on selections of each annotator.}
\end{table*}

Table~\ref{tab:annotation} displays the annotators' selections for each codetype. Additionally, we also record the instances where consensus was achieved among more than two annotators. In cases where consensus among two or more annotators can not be reached, we introduce a fourth annotator for the final decision-making process. Summing up the number of reached consensuses and the decisions made by the fourth annotator yields the final distribution count for codetypes.

To ensure a better understanding of the 10 codetype candidates, we provide corresponding definitions and examples for each candidate. However, due to the diversity of language expressions and potential errors in the original data annotation, we acknowledge the possibility of certain language patterns in the dataset not falling under the 10 specified categories (including \textit{Irony}, \textit{Metaphor}, \textit{Argot}, \textit{Pun}, \textit{Abbreviation}, \textit{Idiom}, \textit{Rhetorical question}, \textit{Loanword}, \textit{Hyperbole}, and \textit{Deformation}). Therefore, we allow annotators to choose \textit{None} as their final response. Nonetheless, we emphasize our preference for annotators to refrain from making such judgments arbitrarily and to strive to assign a codetype to each sentence whenever possible.

\begin{figure}[t]
  \includegraphics[width=\columnwidth]{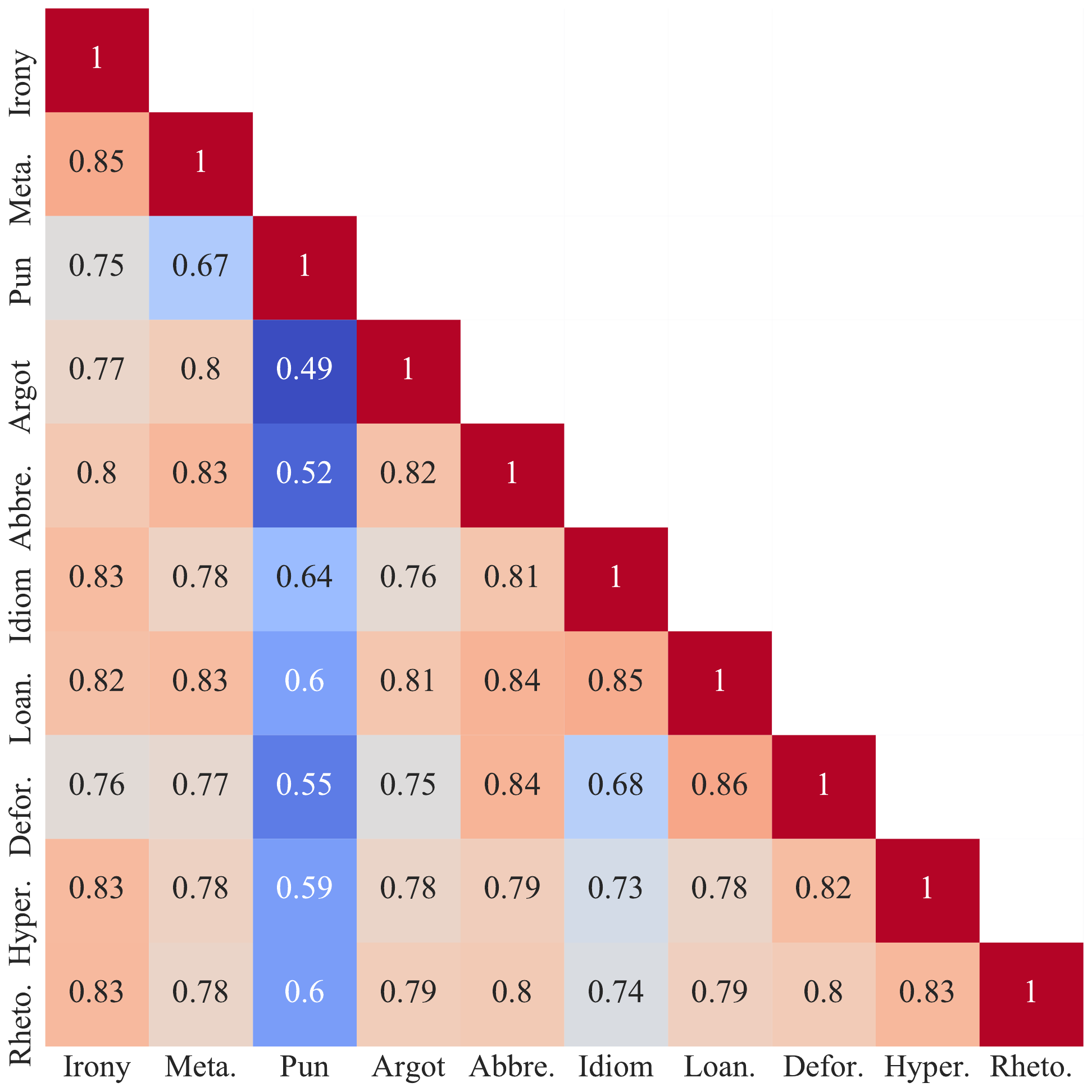}
  \caption{The embedding cosine similarity between 10 codetype candidates using word2vec model.}
   \label{fig:heatmap}
\end{figure}

\subsection{Examples of 10 codetype candidates }\label{app:examples}
Below are the language pattern explanations and 1-2 examples to be referenced during the annotation process. \textit{Examples} of the original Chinese corpus are shown in \textcolor{Melon}{melon} color, with the English translation below it. The encoding keywords related to codetype candidates are \underline{underlined}.

\begin{enumerate}

    \item \textbf{Abbreviation}: In languages using phonetic characters, simplified forms are adopted for commonly used phrases (mostly proper nouns) and a few common words. In Chinese datasets, speakers often use the initial letters of Chinese pinyin for abbreviations. \\
    \textit{Examples}:
    \begin{itemize}
    \item \textcolor{Melon}{\underline{txl}是社会问题，应该有人管，而不是当透明人} \\
       \underline{\pinyin{tong2xing4lian4}} (Homosexual) is a social issue; someone should address it, rather than being invisible.
    \item \textcolor{Melon}{格局小了，国女都是喜欢找黑人\underline{约p}，找国男接盘?} \\
       The scope is limited; Chinese girls all like to find black men for \underline{\pinyin{yue1pao4}} (have casual sex), and Chinese men to take over?
    \end{itemize}

    \item \textbf{Metaphor}: It involves comparing two unrelated things, creating a rhetorical transfer. Speakers liken a certain group to animals or describe a group or object using animal behavior, or they may use specific things or events to describe a group or object. This includes both direct and indirect forms. \\
    \textit{Examples}:
    \begin{itemize}
    \item \textcolor{Melon}{宁愿\underline{小鸡}守活寡？也不想小鸡沾粑粑？} \\
       Would you rather have \underline{a hen} guarding its lone egg than have the hen besmirched?
    \item \textcolor{Melon}{会抱\underline{狗} 但是不会抱\underline{黑人}} \\
       I would hug \underline{a dog} but not \underline{a black person}.
    \end{itemize}

    \item \textbf{Irony}: The literal meaning is inconsistent with the deeper meaning, where words convey something different from their apparent intent. It encompasses both explicit and implicit forms of mockery. Speakers often use words with positive connotations or words of praise to mock individuals or groups. \\
    \textit{Examples}:
    \begin{itemize}
    \item \textcolor{Melon}{不会真有人想和\underline{小仙女}结婚吧不会吧不会吧} \\
       Could anyone really want to marry \underline{little fairy}? Unbelievable.
    \item \textcolor{Melon}{我们同性恋\underline{真是太正确了}} \\
       Our homosexuals are \underline{just too correct}.
    \end{itemize}

    \item \textbf{Pun}: Utilizing homophones or wordplay to convey an indirect meaning. \\
    \textit{Examples}:
    \begin{itemize}
    \item \textcolor{Melon}{\underline{艾}呀，\underline{梅}关系，不会\underline{疣}事的，概率为\underline{淋}。\underline{疣}其湿没有尖锐的问题就更不担心啦，你要照顾好\underline{滋}己呀，不要经常去\underline{疱}吧了，找个\underline{斑}上吧，\underline{疹}心的。} \\
       Oh dear, don't worry, everything will be fine, the probability is zero. Especially with no acute issues, there's even less to worry about. Take care of yourself and don't spend too much time at bars. Go and find a job, sincerely.
    \item \textcolor{Melon}{毕竟同性是真“\underline{艾}”} \\
       Ultimately, being homosexual is true \underline{`LOVE'}.
    \end{itemize}

    \item \textbf{Argot}: Language peculiar to a particular profession or group, incomprehensible to outsiders. In the internet era, internet slang is the jargon among netizens. Different platforms have different popular expressions. Examples of derogatory internet slang commonly seen on Zhihu and Tieba include: 拳 (boxer) refers to someone who holds sexist views, 被绿 (getting greened) refers to being cheated on in a relationship, 妈宝女 (mommy's girl) or 妈宝男 (mommy's boy) are terms used to describe adult males or females who obediently follow their mothers' wishes, often lacking independent thinking and decision-making abilities. \\
    \textit{Examples}:
    \begin{itemize}
    \item \textcolor{Melon}{娶媳妇没用的关键是大小不然会\underline{被绿}当老实人接盘侠，这么多年没有任何改变} \\
       The key to marrying a daughter-in-law is to have status; otherwise, you'll \underline{be cheated}, regarded as an honest man taking over the role. After so many years, there's been no change at all.
    \item \textcolor{Melon}{\underline{妈宝女}：我妈说彩礼三十万} \\
       \underline{Mommy's girl}: My mom says the dowry should be thirty thousand.
    \end{itemize}

    \item \textbf{Idiom}: Informal and colloquial words or phrases used by the general public, also known as dialect, vernacular, or folk idioms, including vocabulary or short phrases. For instance, 基佬 (gay) originally appeared as a dialect in the Guangdong region. \\
    \textit{Examples}:
    \begin{itemize}
    \item \textcolor{Melon}{东北人自强不息就OK了，这么肥沃的黑土，丰富的矿藏。不发达还是要自己找找原因是不是太大方了?毕竟中原有句老话叫做\underline{斗米恩，升米仇}} \\
       Northeasterners should strive on their own. With such fertile black soil and abundant mineral resources, why are they still underdeveloped? Could it be because they're too generous? After all, there's an old saying in the Central Plains that goes \underline{Feud rice is fought, rise rice is resented}.
    \item \textcolor{Melon}{\underline{苍蝇不叮无缝蛋，自食恶果}} \\
       \underline{Flies don't sting seamless eggs; they re-} \underline{ap what they sow.}
    \end{itemize}

    \item \textbf{Loanword}: Words borrowed from another language. Generally, loanwords do not include translated words but refer to words that are absorbed with both sound and meaning. For example, 凯勒奇 (Kaleqi) is the English translation of the anti-Semitic Kalergi Plan. \\
    \textit{Examples}:
    \begin{itemize}
    \item \textcolor{Melon}{犹太人正在中国借助女权LGBT推动黑人\underline{凯乐其计划}} \\
       Jews are promoting the black \underline{Kalergi Plan} in China through feminism and LGBT rights.
    \item \textcolor{Melon}{再次证实了\underline{凯乐奇计划}的现实。} \\
       Once again, it proves the reality of the \underline{Kalergi Plan}.
    \end{itemize}

    \item \textbf{Decomposition}: As Chinese characters are ideograms, they can convey specific emotions through individual characters' separation and combination. For example, The character `默'(silence) is composed of `黑' (black) + `犬' (dog), which is used to mock black people. \\
    \textit{Examples}:
    \begin{itemize}
    \item \textcolor{Melon}{好奇纹了什么字，\underline{默}吗} \\
       I wonder what character was used to create curiosity, \underline{`默'}?
    \end{itemize}

    \item \textbf{Hyperbole}: Intentionally magnifying and embellishing the characteristics of objective persons, events, or things to deviate from the truth, aiming to deepen the reader's impression.\\
    \textit{Examples}:
    \begin{itemize}
    \item \textcolor{Melon}{我觉得，你可以在淘宝开定制戒指服务，接单给小黑做，你这等于有一个\underline{几百个小工的工厂}啊} \\
       I think you could offer custom ring services on Taobao, and take orders for Little Black People, it's like having a \underline{factory with hundreds of workers}.
    \end{itemize}

    \item \textbf{Rhetorical Question}: The speaker poses a question that seems to be directed at the reader or audience, but actually contains the author's own answer. This answer may be explicit or implicit. This technique is mainly used to emphasize viewpoints, guide thinking, or evoke emotions.
    
    \textit{Examples}:
    \begin{itemize}
    \item \textcolor{Melon}{四川的黑点\underline{不是}gay多\underline{嘛}?} \\
       \underline{Isn't} Sichuan full of gays?
    \end{itemize} 
\end{enumerate}

\section{More details about codetype strategy}

\subsection{Prompt-based method}
\label{app:prompt-based methd}
We use the exact same prompts for both English and Chinese datasets. We have shown the user prompt in Section~\ref{sec:prompt-based method}. The system prompt we utilize here is as follows:

\begin{prompt}[title={\textbf{System prompt}}]
Please answer the question strictly according to the given instructions.
\end{prompt}

\end{CJK*}
\end{document}